\documentclass{jpsj3}
\usepackage{graphicx}
\usepackage{mathbbol}
\usepackage{txfonts}
\usepackage{color}
\title{Statistical-mechanical analysis of pre-training and fine tuning in deep learning}
\author{Masayuki Ohzeki \thanks{mohzeki@i.kyoto-u.ac.jp}}
\inst{Department of Systems Science, Graduate School of Informatics, Kyoto University, 36-1 Yoshida Honmachi, Sakyo-ku, Kyoto, 606-8501, Japan} 
\abst{
In this paper, we present a statistical-mechanical analysis of deep learning. We elucidate some of the essential components of deep learning---pre-training by unsupervised learning and fine tuning by supervised learning. We formulate the extraction of features from the training data as a margin criterion in a high-dimensional feature-vector space. The self-organized classifier is then supplied with small amounts of labelled data, as in deep learning. Although we employ a simple single-layer perceptron model, rather than directly analyzing a multi-layer neural network, we find a nontrivial phase transition that is dependent on the number of unlabelled data in the generalization error of the resultant classifier. In this sense, we evaluate the efficacy of the unsupervised learning component of deep learning. The analysis is performed by the replica method, which is a sophisticated tool in statistical mechanics. We validate our result in the manner of deep learning, using a simple iterative algorithm to learn the weight vector on the basis of belief propagation.}

\begin{document}

\maketitle

\section{\protect\normalsize Introduction}
Deep learning is a promising technique in the field of machine learning, with its outstanding performance in pattern recognition applications, in particular, being extensively reported. The aim of deep learning is to efficiently extract important structural information directly from the training data to produce a high-precision classifier \cite{Hinton2006}.
The technique essentially consists of three parts. First, a large number of hidden units are introduced by constructing a multi-layer neural network, known as a deep neural network (DNN). This allows the implementation of an iterative coarse-grained procedure, whereby each high-level layer of the neural network extracts abstract information from the input data. In other words, we introduce some redundancy for feature extraction and dimensional reduction (a kind of sparse representation) of the given data.
The second part is pre-training by unsupervised learning.
This is a kind of self-organization \cite{Kohonen2001}.
To accomplish self-organization in the DNN, we provide plenty of unlabelled data. The network learns the structure of the input data by tuning the weight vectors (often termed the network parameters) assigned to each layer of the neural network. The procedure of updating each weight vector on the basis of the gradient method, i.e., back propagation, takes a relatively long time \cite{Bishop2006} and its regularization by $L_1$ norm and greedy algorithm \cite{Sohl-Dickstein2011,Aurelien2014,Yamanaka2015}.
This is because many local minima are found during the optimization of the DNN. Instead, techniques such as the auto-encoder have been proposed to make the pre-training more efficient and push up the basins of attraction of the minima via a better generalization of the training data \cite{Hinton2006sci,Bengio2009,Erhan2010}.
The third component of deep learning involves fine tuning the weight vectors using supervised learning to elaborate DNN into a highly precise classifier. This combination of unsupervised and supervised learning enables the architecture of deep learning to obtain better generalization, effectively improving the classification under a semi-supervised learning approach \cite{Seeger2002,Zhu2008}.

In the present study, we focus on the latter two parts of deep learning. 
The first is neglected because it simply highlights a way of implementing the deep learning algorithm. 
A recent study has formulated a theoretical basis for the relationship between the recursive manipulation of variational renormalization groups and the multi-layer neural network in deep learning \cite{RG2014}.
Indeed, it is confirmed that the renormalization group indeed can mitigate the computational cost in the learning without any significant degradation \cite{Tanaka2015}.
Furthermore, the direct evaluation of multi-layer neural networks is too complex to fully clarify the early stages of our theoretical understanding of deep learning. 
Although most of the DNN is constructed by a Boltzmann machine with hidden units, we simplify the DNN to a basic perceptron. 
This simplification, which is just for our analysis, enables us to shed light on the fundamental origin of the outstanding performance of deep learning and the efficiency of pre-training by unsupervised learning.

The steady performance of the classifier constructed by the deep learning algorithm can be assessed in terms of the generalization error using a statistical-mechanical analysis based on the replica method \cite{Nishimori2001}.
We consequently find nontrivial behaviour involved in the emergence of the metastable state of the generalization error, a result of the combination of unsupervised and supervised learning. This is analogous to the metastable state in classical spin models, which leads to the hysteresis effect in magnetic fields. Following the actual process of deep learning, we numerically test our result by successively implementing the unsupervised learning of the pre-training procedure and the supervised learning for fine tuning. We then demonstrate the effect of being trapped in the metastable state, which worsens the generalization error. This justifies the need for fine tuning by several sets of labelled data after the pre-training stage of deep learning.

The remainder of this paper is organized as follows. In the next section, we formulate our simplified model to represent unsupervised and supervised learning with structured data, and analyze the Bayesian inference process for the weight vectors. In Section 3, we investigate the nontrivial behaviour of the generalization error in our model. We demonstrate that the generalization error can be significantly improved by the use of sufficient amounts of unlabelled data. Finally, in Section 4, we summarize the present work.
\section{Analysis of combination of unsupervised and supervised learning}
\subsection{\protect\normalsize Problem setting}
We deal with a simple two-class labelled-unlabelled classification problem.
We assume that the $N$-dimensional feature vectors ${\bf x}_{\mu} \in \mathbb{R}^N$ obey the following distribution function conditioned on the binary label $y_{\mu} = \pm 1$ for each datum $\mu$ and a predetermined weight vector ${\bf w}_0$:
\begin{equation}
P_g({\bf x}_{\mu}|y_{\mu},{\bf w}_0) \propto \Theta\left( \frac{y_{\mu}}{\sqrt{N}}{\bf x}_{\mu}^{\rm T}{\bf w}_0 - g\right),
\end{equation}
where $g$ is a margin, which resembles the structure of the feature vectors of the given data, and
\begin{equation}
\Theta(x) = \left\{
\begin{array}{cc}
1 & x > 0 \\
0 & x \le 0
\end{array}
\right. .
\end{equation}
The labelled data $({\bf x}_{\mu},y_{\mu})$ ($\mu = 1,2,\cdots,L$) are generated from the joint probability $P_g({\bf x}_{\mu}|y_{\mu},{\bf w}_0)P(y_{\mu})$, where $L$ is the number of labelled data.
The unlabelled data $({\bf x}_{\mu})$ ($\mu = L+1,L+2,\cdots,L+U$), where $U$ is the number of unlabelled data, follow the marginal probability $P_g({\bf x}_{\mu}|{\bf w}_0) = \sum_{y_{\mu}} P_g({\bf x}_{\mu}|y_{\mu},{\bf w}_0)P(y_{\mu})$.
In the following, we assume the large-$N$ limit and a huge number of data $L,U\sim O(N)$, as well as a  symmetric distribution for the label $P(y_{\mu}) = 1/2$.

The likelihood function for the dataset is defined as
\begin{equation}
P_g(\mathcal{D}|{\bf w}_0) = \prod_{\mu = 1}^L P_g({\bf x}_{\mu}|y_{\mu},{\bf w}_0)P(y_{\mu}) \prod_{\mu = L+1}^{L+U} P_g({\bf x}_{\mu}|{\bf w}_0),\label{LF}
\end{equation}
where $\mathcal{D}$ denotes the dataset consisting of labelled data and unlabelled data.
When the  feature vector $g$ has a margin value of zero, unsupervised learning is no longer meaningful, because the marginal distribution becomes flat.
However, nonzero values of the margin  elucidate the structure of the feature vectors through the unsupervised learning.
The actual data in images and sounds have many inherent structures that must be represented by high-dimensional weight vectors in the multi-layer neural networks of DNN.
In the present study, we simplify this aspect of the actual data to give an artificial model with a margin that follows the simple perceptron. This allows us to assess certain nontrivial aspects of deep learning.

\subsection{Bayesian inference and replica method}
For readers unfamiliar with deep learning, we sketch the procedure of the deep learning here. 
The first step of the deep learning algorithm is to conduct pre-training.
Following the unsupervised learning, the weight vector learns the features of the training data without any labels.
As a simple strategy, we often estimate the weight vector to maximize the likelihood function only for the unlabelled data as
\begin{equation}
{\bf w}^{\rm PT} = \arg \max_{\bf w} \left\{ \log \prod_{\mu = L+1}^{L+U} P_h({\bf x}_{\mu}|{\bf w}) \right\}.
\end{equation}
We use a different margin value $h$ from one in Eq. (\ref{LF}) in order to evaluate a generic case below.
When we know a priori the structure of the data, one may set $g=h$.
We may utilize the hidden units to prepare some redundancy to represent the feature of the given data.
In the present study, we omit this aspect to simplify the following analysis.
In other words, we have a coarse-graining picture of DNN only by a single layer with a weight vector ${\bf w}$, the input ${\bf x}_{\mu}$ and output $y_{\mu}$.
In the second step, termed as the fine tuning step, we estimate the weight vector to precisely classify the training data.
For instance, the maximum likelihood estimation can be a candidate to estimate the weight vector as 
\begin{equation}
{\bf w}^{\rm FT} = \arg \max_{\bf w} \left\{ \log  \prod_{\mu = 1}^L P_h({\bf x}_{\mu}|y_{\mu},{\bf w})P(y_{\mu}) \prod_{\mu = L+1}^{L+U} P_h({\bf x}_{\mu}|{\bf w}) \right\}.\label{FT}
\end{equation}
We notice an important thing of the deep learning architecture.
In this procedure, we use the result of the pre-training ${\bf w}^{\rm PT}$ as an initial condition for the gradient method to obtain ${\bf w}^{\rm FT}$.
The purpose of the deep learning is just obtain the weight vector to classify the newly-generated data with better performance simply from some strategy as in Eq. (\ref{FT}).
The computational cost of the often-employed methods (e.g. back propagation \cite{Bishop2006}) becomes extremely longer in general.
However if we have some adequate initial condition to manipulate the estimation, we can mitigate harmful computation and reach a better estimation of the weight vector \cite{Bengio2009,Erhan2010}. 

In order to evaluate the theoretical limitation of the deep learning, instead of the maximum likelihood estimation, we employ an optimal procedure based on the framework of Bayesian inference.
The posterior distribution can be given by the Bayes' formula as
\begin{equation}
P_h({\bf w}|\mathcal{D}) = \frac{P_h(\mathcal{D}|{\bf w})P({\bf w})}{\int d{\bf w}'P_h(\mathcal{D}|{\bf w}')P({\bf w}')}.
\end{equation}
We assume that the prior distribution for the weight vector is $P({\bf w}) \propto \delta\left(|{\bf w}|^2 - N\right)$.
The posterior mean given by this posterior distribution provides an estimator for the quantity related to the weight vector:
\begin{equation}
\mathbb{E}_{{\bf w}|\mathcal{D}}[f({\bf w})] = \int d{\bf w}f({\bf w})\frac{P_h(\mathcal{D}|{\bf w})P({\bf w})}{\int d{\bf w}'P_h(\mathcal{D}|{\bf w}')P({\bf w}')}.
\end{equation}
The typical value is evaluated by averaging over the randomness of the dataset as
\begin{equation}
\mathbb{E}_{\mathcal{D}}[ \mathbb{E}_{{\bf w}|\mathcal{D}}[g({\bf w})]] = \left[ \int d{\bf w}g({\bf w})\frac{P_h(\mathcal{D}|{\bf w})P({\bf w})}{\int d{\bf w}'P_h(\mathcal{D}|{\bf w}')P({\bf w}')}\right]_{\mathcal{D}},
\end{equation}
where
\begin{equation}
[\cdots]_{\mathcal{D}} = \int d\mathcal{D} d{\bf w}_0 P_g(\mathcal{D}|{\bf w}_0)P({\bf w}_0) \times \cdots.
\end{equation}
The average quantity is given by the derivative of the characteristic function, namely the free energy, which is defined as
\begin{eqnarray}
- \mathcal{F} = \lim_{N \to \infty}\frac{1}{N}\left[ \log \int d{\bf w} P_h(\mathcal{D}|{\bf w})P({\bf w}) \right]_{\mathcal{D}}.
\end{eqnarray}
In particular, as shown below, the derivative of the free energy yields a kind of self-consistent equations for the physically-relevant quantities.
In this problem, we compute the overlap between the estimated ${\bf w}$ and the original weight vectors ${\bf w}_0$ and the variance of the weight vectors, which quantify the precision of the learning.
Following spin glass theory \cite{Nishimori2001}, we apply the replica method to evaluate the free energy.
We define the replicated partition function as
\begin{equation}
\Xi_n = \left( \int d{\bf w} P_h(\mathcal{D}|{\bf w})P({\bf w})\right)^n .
\end{equation}
The (density of) free energy can be calculated from the replicated partition function through the replica method as
\begin{eqnarray}
- \mathcal{F} &=&  \lim_{n \to 0} \frac{\partial}{\partial n} \lim_{N \to \infty}\frac{1}{N}\log\left[  \Xi_n \right]_{\mathcal{D}}.
\end{eqnarray}
We exchange the order of the operations on $n$ and the thermodynamic limit $N \to \infty$, and assume that the replica number $n$ is temporarily a natural number in the evaluation of $[\Xi_n]_{\mathcal{D}}$.
We introduce the following constraints to simplify the calculation dependent on ${\bf w}_a$:
\begin{equation}
\int dQ \prod_{a \ge b} \delta\left( Q_{ab} - \frac{1}{N}{\bf w}_a{\bf w}_b\right)\prod_{a=0} \delta\left( Q_{0a} - \frac{1}{N}{\bf w}_0{\bf w}_a\right).
\end{equation}
The free energy is then given by solving an extremization problem:
\begin{equation}
- \mathcal{F} = \sup_{Q} \left[ \mathcal{G}(Q) - \mathcal{I}(Q) \right],\label{FE1}
\end{equation}
where
\begin{eqnarray}\nonumber
\mathcal{G}(Q) &=& \alpha \log \left[ \Theta \left( u_{0} - g \right) \prod_{a=1}^n \Theta \left( u_{\alpha} - h \right) \right]_{\bf u} + \beta \log \left[ \Phi(u_0,g) \prod_{a=1}^n \Phi(u_a,h) \right]_{\bf u} \\
\\
\mathcal{I}(Q) &=& \sup_{\tilde{Q}}\left( \sum_{a \ge b} Q_{ab}\tilde{Q}_{ab} + \sum_{a=1}^nQ_{0a}\tilde{Q}_{0a} - \log \mathcal{M}(\tilde{Q})\right)\\
\mathcal{M}(\tilde{Q}) &=& \mathbb{E}_{{\bf w}} \left[ \exp\left( \sum_{a \ge b} \tilde{Q}_{ab}{\bf w}_a{\bf w}_b + \sum_{a=1}^n\tilde{Q}_{0a}{\bf w}_0{\bf w}_a \right)\right].
\end{eqnarray}
Here, $\alpha = L/N$, $\beta = U/N$, and
\begin{equation}
\Phi(u,h) = \frac{1}{2}\Theta \left( u - h \right) + \frac{1}{2}\Theta \left( -u - h \right).
\end{equation}
The expectation is taken over the distribution $\prod_{a=0}^nP({\bf w}_a)$.
We introduce auxiliary parameters $\tilde{Q}_{ab}$ to give an integral representation of the Kronecker's delta.
We use $[\cdots]_{\bf u}$ to denote the average with respect to the $(n+1)$-multivariate Gaussian random variables $\{u_a\}$ with  vanishing mean and  covariance $[u_au_b]_{\bf u} = \delta_{ab}+Q_{ab}(1-\delta_{ab})$.

\subsection{Replica-symmetric solution}
Let us evaluate the replica-symmetric solution by imposing invariant symmetry for $Q_{ab}$ and $\tilde{Q}_{ab}$ under permutation of the replica index as
\begin{equation}
\begin{array}{ccc}
Q_{aa} = 1 & Q_{ab} = q & Q_{0a} = m \\
\tilde{Q}_{aa} =\tilde{Q} & \tilde{Q}_{ab} = \tilde{q} & \tilde{Q}_{0a} = \tilde{m}.
\end{array}
\end{equation}
Then, the Gaussian random variables can be written as $u_{a} = \sqrt{q}z+\sqrt{1-q}t_{a}$ for $a>0$ and $u_{0} = \sqrt{m^2/q}z+\sqrt{1-m^2/q}t_0$ using the auxiliary normal Gaussian random variables $\{t_a\}$ and $z$ with vanishing mean and unit variance.
Under the RS assumption, we obtain an explicit form for the free energy by solving the saddle-point equation for $\tilde{Q}$, $\tilde{q}$, and $\tilde{m}$:
\begin{eqnarray}\nonumber
-\mathcal{F}&=& \alpha\int Dz H\left(\frac{mz+\sqrt{q}g}{\sqrt{q-m^2}}\right) \log H \left(\frac{\sqrt{q}z+h}{\sqrt{1-q}}\right) \\ 
&&  +\beta\int Dz G_g(m,\sqrt{q}) \log G_h(\sqrt{q},1) + \frac{1}{2} \log(1-q) + \frac{q-m^2}{2(1-q)}\label{freeE},
\end{eqnarray}
where $Dz=dz\exp(-z^2/2)$, $H(x) = \int^{\infty}_x Dt$, and
\begin{eqnarray}
G_h(a,b) &=& \frac{1}{2}\left\{ H\left(\frac{az+bh}{\sqrt{b^2-a^2}}\right) + H\left(\frac{az-bh}{\sqrt{b^2-a^2}}\right)\right\}.
\end{eqnarray}
The partial derivatives of the free energy (\ref{freeE}) with respect to  $m$ and $q$ lead to the saddle-point equations for the physically-relevant RS order parameters, namely the overlap $m$ and the variance $w$ of the weight vector:
\begin{eqnarray}\nonumber
&&\alpha \int Dz  H'\left(\frac{mz+\sqrt{q}g}{\sqrt{q-m^2}}\right) \left( \frac{ H'\left(\frac{\sqrt{q}z+h}{\sqrt{1-q}}\right) }{H \left(\frac{\sqrt{q}z+h}{\sqrt{1-q}}\right)}\right) \\ && + \beta \int Dz  G'_g(m,\sqrt{q}) \left( \frac{ G'_h(\sqrt{q},1)}{G_h(\sqrt{q},1)}\right) = \frac{m}{1-q}, \label{SP1}\\ \nonumber
&&\alpha \int Dz   H\left(\frac{mz+\sqrt{q}g}{\sqrt{q-m^2}}\right) \left( \frac{ H'\left(\frac{\sqrt{q}z+h}{\sqrt{1-q}}\right) }{H \left(\frac{\sqrt{q}z+h}{\sqrt{1-q}}\right)}\right)^2 \\ && + \beta \int Dz  G_g(m,\sqrt{q}) \left( \frac{ G'_h(\sqrt{q},1)}{G_h(\sqrt{q},1)}\right)^2 =\frac{q-m^2}{(1-q)^2}, \label{SP2}
\end{eqnarray}
where $H'(x) = - \exp(-z^2/2)/\sqrt{2\pi}$ and
\begin{eqnarray}
G'_h(a,b) &=& \frac{1}{2}\left\{ H'\left(\frac{az+bh}{\sqrt{b^2-a^2}}\right) - H'\left(\frac{az-bh}{\sqrt{b^2-a^2}}\right)\right\}.
\end{eqnarray}
The RS solution always satisfies $q=m$ under the condition $g = h$ (the Bayes-optimal solution).
The above saddle-point equations are then reduced to the following single equation for $q$:
\begin{eqnarray}
&&\alpha \int Dz   \frac{\left( H'\left(\frac{\sqrt{q}z+h}{\sqrt{1-q}}\right) \right)^2}{H\left(\frac{\sqrt{q}z+h}{\sqrt{1-q}}\right)} + \beta \int Dz \frac{\left( G'_h(\sqrt{q},1) \right)^2}{G_h(\sqrt{q},1)} =\frac{q}{1- q}.
\label{SPeq}
\end{eqnarray}
The order parameter $q$ is closely related to the generalization error, which is defined as the probability of disagreement between the labelled data and the classifier outputs for the newly generated example after the classifier has been trained.
In the case of an input--output relation given by a simple perceptron, the generalization error is expressed as \cite{Nishimori2001}:
\begin{equation}
\epsilon = \frac{1}{\pi} \cos^{-1}q.
\end{equation}
We will evaluate this quantity to validate the performance of the classifier generated from the combination of unsupervised  and supervised learning.

\section{Saddle point and numerical verification}
\begin{figure}[tb]
\begin{center}
\includegraphics[width=160mm]{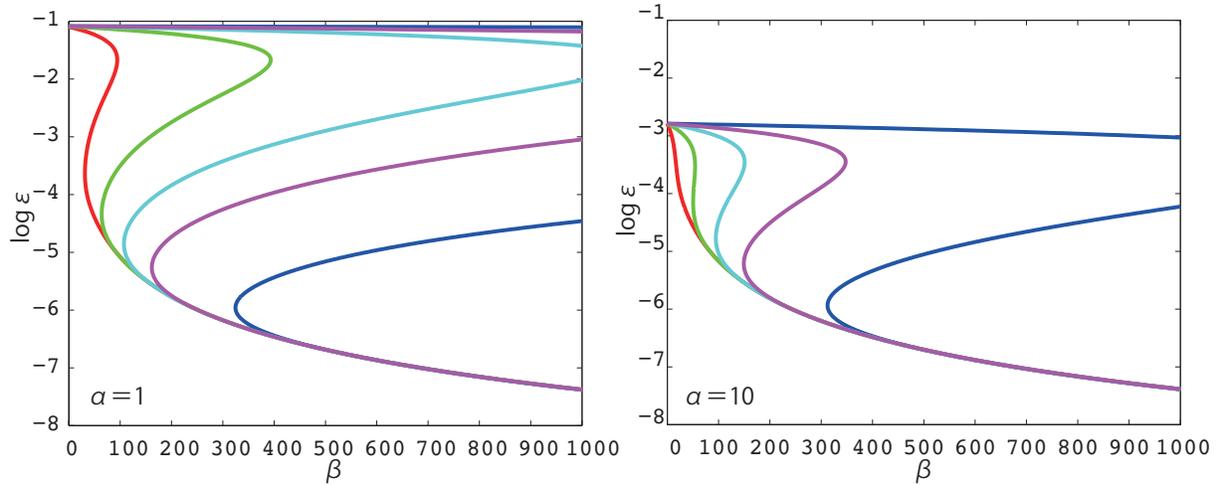}
\end{center}
\caption{{\protect\small (color online) Generalization errors for $h=0.1, 0.05, 0.03, 0.02$, and $0.01$ (curves from left to right).
The left panel shows the results for $\alpha = 1$, and the right one represents $\alpha=10$.
Both cases exhibit multiple solutions for the same value of $\beta$.
}}
\label{fig1}
\end{figure}

In Fig.~\ref{fig1}, we plot the logarithm of the generalization error  with respect to the number of supervised learning data for several values of $h$. Each plot shows the results for a different value of $\alpha$.
Note that when there is no fine tuning through supervised learning (i.e., $\alpha = 0$), the generalization error does not exhibit any nontrivial behaviour.
However, for nonzero $\alpha$, we find nontrivial curves, which give multiple solutions for the same $\beta$, in the $\beta-\epsilon$ plane.
This is a remarkable result for the combination of unsupervised and supervised learning.
The nontrivial curves imply the existence of a metastable state, similar to several classical spin models \cite{Nishimori2011}.
As $h$ decreases, the spinodal point $\beta_{\rm sp}$ (the point at which  the multiple solutions coalesce) moves to larger values of $\beta$.
This is because decreasing $h$ leads to difficulties in the classification of the input data.
In other words, we need a vast number of unlabelled data to attain the lower-error state for a fixed number of labelled data.
However, the metastable state remains up to a large value of $\beta$, causing the computational cost to become very expensive.
We therefore need an extremely long computational time to reach the lower-error solution, or find good initial conditions nearby.
On the other hand, increasing $\alpha$ causes the spinodal points to move to lower values of $\beta$.
Although this confirms an improvement in the generalization error for the higher-error state, there is no quantitative change in that for the lower-error state.
In this sense, pre-training is an essential part of the architecture of deep learning if we wish to achieve the lower-error state---this is the origin of deep learning's remarkable performance.
In contrast, the emergence of the metastable state causes the computational cost to increase drastically.
Several special techniques could be incorporated into the architecture of deep learning to  avoid this weak point, effectively preparing good initial conditions that enable the lower-error state to be reached \cite{Bengio2009,Erhan2010}.

The asymptotic form of $H(x) \sim \Theta(x)\exp(-x^2/2)/|x|$ for $x \to \infty$ leads to the exponent of the learner curve \cite{Nishimori2001}, which characterizes the decrease in the generalized error in $\alpha \gg 1$ and $\beta \gg 1$ as $\epsilon_g \sim (c_{\alpha}\alpha^2 + c \alpha \beta +c_{\beta}\beta^2)^{-1}$.
Here, $c_{\alpha}$, $c_{\beta}$, and $c$ are the constants evaluated by the Gaussian integrals.
Thus, there is no quantitative change in the exponent of the learning curve in this formulation compared with  that of the perceptron with ordinary supervised learning.

Next, let us consider the effect of fine tuning in the context of deep learning.
If we plot the saddle-point solutions in the $\alpha-\epsilon$ plane,
we find that  multiple solutions appear in a certain region.
Increasing the number of unlabelled data again leads to an improvement in the generalization error.
A gradual increase in the number of labelled data allows us to escape from the metastable state.
In this sense, fine tuning by supervised learning is necessary to achieve the lower-error state and mitigate the difficulties in reaching the desired solution.
We should emphasize that the emergence of the metastable state does not come  from the multi-layer neural networks in DNN, but from the combination of unsupervised and supervised learning.
This observation was also noted in a previous study \cite{Tanaka2013}.

\begin{figure}[tb]
\begin{center}
\includegraphics[width=100mm]{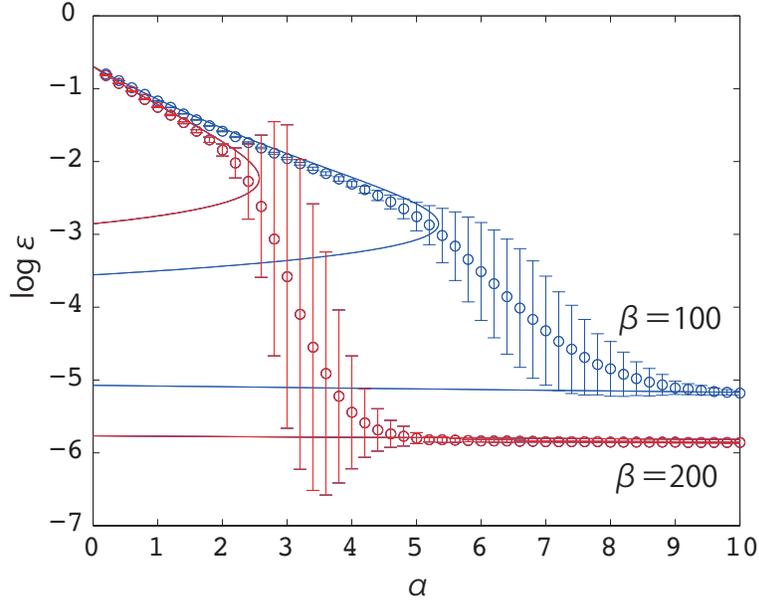}
\end{center}
\caption{{\protect\small (Color online) Numerical test using approximate message passing.
We illustrate the case with $h=0.05$ for $\beta=100$ (blue) and $\beta=200$ (red).
Error bars are shown for each plot over $N_{\rm sam}=1000$ samples.
}}
\label{fig2}
\end{figure}

To verify our analysis, we conduct numerical experiments using the so-called approximate message passing algorithm \cite{Kabashima2003}.
On the basis of the reference in the modern fashion \cite{YingYing2014}, we can construct an iterative algorithm to infer the weight vector using both the unlabelled and labelled data.
The update equations are
\begin{eqnarray}
a^{t+1}_{\mu} &=& \sum_{k=1}^N x_{\mu k} w_k - \frac{1}{\kappa^t}C_{\mu}\left(a^t_{\mu},\frac{1}{\kappa^t},h\right) \\
\kappa^{t+1} &=& \frac{1}{2}\left( 1 + \frac{4}{N}\sum_{k=1}^N\left(a^t_k\right)^2\right)\\
a^{t+1}_k &=& \sum_{\mu=1}^{L+U} x_{\mu k} C_{\mu}\left(a^t_{\mu},\frac{1}{\kappa^t},h\right) + B^{t} a^{t}_k \\
B^{t+1} &=& \frac{1}{N} \sum_{\mu=1}^{L+U}D_{\mu}\left(a^t_{\mu},\frac{1}{\kappa^t},h\right),
\end{eqnarray}
where
\begin{eqnarray}
C_{\mu}(a,b,h) &=&
\left\{
\begin{array}{cc}
y_{\mu} \displaystyle\frac{\exp(-z_-^2/2)}{\sqrt{2\pi b}H(z_-)} & (\mu \le L) \\
\displaystyle\frac{\exp(-z_-^2/2)-\exp(-z_+^2/2)}{\sqrt{2\pi b}(H(z_-) + H(z_+))} & (\mu > L)
\end{array}
\right.\\ 
D_{\mu}(a,b,h) &=&
\left\{
\begin{array}{cc}
C^2_{\mu}(a,b,h)
 - y_{\mu} \displaystyle\frac{z_-C_{\mu}(a,b,h)}{\sqrt{b}} & (\mu \le L) \\
C^2_{\mu}(a,b,h) + \displaystyle\frac{z_-\exp(-z_-^2/2)+z_+\exp(-z_+^2/2)}{\sqrt{2\pi}b(H(z_-) + H(z_+))} & (\mu > L).
\end{array}
\right. 
\end{eqnarray}
Here $x_{\mu k}$ is the $k$th component of the feature vector of the datum $\mu$ and $w_k$ is the $k$th component of the weight vector.
We use the abbreviation $z_{\pm} = (h \pm a)/\sqrt{b}$,
and estimate the weight vector from ${\bf w} = {\bf a}^{t}/\kappa^t$.
In the numerical experiments, we first estimate the weight vector using only the unlabelled data, i.e., $\alpha = 0$.
We then  gradually increase the number of labelled data while estimating the weight vector.
The system size is set to $N=100$, and the number of samples $N_{\rm sam} = 1000$.
The maximum iteration number for fine tuning is set to $20$.
In Fig.~\ref{fig2}, we plot the average generalization error over $N_{\rm sam}$ independent runs starting from the randomized initial conditions.
As theoretically predicted, our results confirm the water-falling phenomena for several cases with $h=0.5$.
Increasing the number of labelled data in the fine tuning step allows us to escape from the metastable state.
Therefore, fine tuning is a necessary component in the remarkable performance of deep learning.
However, the difficulty of classification, represented by $h$, demands a large number of training data.
Therefore, we require the initial condition to be as good as possible in the fine tuning to reach the lower-error state.
Several empirical studies of the deep learning algorithm have revealed that special techniques such as the auto-encoder can provide  initial conditions that are sufficiently good to improve the performance after fine tuning \cite{Erhan2010}.
In future work, we intend to clarify that such specific techniques do indeed overcome the degradation in performance caused by the metastable state.

\section{Conclusion}
We have analyzed the simplified perceptron model under a combination of unsupervised and supervised learning for data with a margin.
The margin imitates the structure of the training data.
We have found nontrivial behaviour in the generalization error of the classifier obtained by this hybrid of unsupervised and supervised learning.
First, we confirmed the remarkable improvement in the generalization error by increasing the number of unlabelled data.
In this sense, the pre-training step in  deep learning is essential when  few labelled data are available.
In addition, our result reveals the existence of the metastable solution, which hampers the ordinary gradient-based iteration to pursue the optimal estimation.
In the deep learning algorithm, the pre-training technique is crucial in reducing the computation time and attaining good performance, because good initial conditions allow the algorithm to reach the lower-error state.
Instead of focusing on the specialized  pre-training technique, we have investigated a nontrivial behaviour involved in the metastable state and the existence of the lower-error state, which is used in the deep learning.
In addition, we have analyzed the role of fine tuning by changing the number of labelled data.
This also confirmed the nontrivial behaviour in the generalization error.
Our numerical experiments demonstrated the water-falling phenomena involved in the existence of the metastable state and confirms that after fine tuning we reach the lower-error state.

We make a remark on the statistical-mechanical analysis for a similar problem setting, namely that of semi-supervised learning.
A previous analysis also revealed the existence of the metastable state \cite{Tanaka2013}.
The present study suggests that the metastable state is essential in the combination of unsupervised and supervised learning.
In this sense, for the sake of the further development to efficiently perform the deep learning, we should invent some techniques to escape from the metastable state, 

Our present work is one instance in which a simplified model can demonstrate the essence of deep learning and clarify certain theoretical aspects.
We hope that future studies will ``extract the features'' of the architecture of deep learning.
\section*{Acknowledgments}
The author would like to thank  Muneki Yasuda, Jun-ichi Inoue, and Tomoyuki Obuchi for fruitful discussions.
The present work has been performed with financial support from the JST-CREST, MEXT KAKENHI Grant Nos. 25120008 and 24740263, and the Kayamori Foundation of Informational Science Advancement.
\bibliography{67003}
\end{document}